\documentclass[10pt,twocolumn,letterpaper]{article}
\makeatletter
\providecommand{\@LN}[2]{}
\makeatother
\usepackage[pagenumbers]{./cvpr} 

\definecolor{cvprblue}{rgb}{0.21,0.49,0.74}
\usepackage[pagebackref,breaklinks,colorlinks,allcolors=cvprblue]{hyperref}
\usepackage{multirow} 
\usepackage{colortbl}
\usepackage{float}
\usepackage[linesnumbered, ruled]{algorithm2e}

\SetCommentSty{mycommfont}
\def\paperID{15753} 
\def\confName{CVPR}
\def\confYear{2025}

\title{Towards Fine-Grained Interpretability: \\Counterfactual Explanations for Misclassification with Saliency Partition}

\newcommand{\authorskip}{\hspace{10.0mm}}

\author{
Lintong Zhang\thanks{These authors contributed equally to this work.} \authorskip 
Kang Yin\footnotemark[1] \authorskip 
Seong-Whan Lee\thanks{Corresponding author.} \\[2.5pt]
\normalsize Dept. of Artificial Intelligence, Korea University, Seoul, Korea\\[-1pt]
{\tt\small \{zhanglintong, charles\_kang, sw.lee\}@korea.ac.kr}\\
}

\begin{document}
\maketitle
\begin{abstract}

Attribution-based explanation techniques capture key patterns to enhance visual interpretability; however, these patterns often lack the granularity needed for insight in fine-grained tasks, particularly in cases of model misclassification, where explanations may be insufficiently detailed. To address this limitation, we propose a fine-grained counterfactual explanation framework that generates both object-level and part-level interpretability, addressing two fundamental questions: (1) which fine-grained features contribute to model misclassification, and (2) where dominant local features influence counterfactual adjustments. Our approach yields explainable counterfactuals in a non-generative manner by quantifying similarity and weighting component contributions within regions of interest between correctly classified and misclassified samples. Furthermore, we introduce a saliency partition module grounded in Shapley value contributions, isolating features with region-specific relevance. Extensive experiments demonstrate the superiority of our approach in capturing more granular, intuitively meaningful regions, surpassing fine-grained methods.
 
\end{abstract}    
\section{Introduction}
\label{sec:intro}
Deep learning systems have significantly advanced computer vision; however, their "black-box" nature remains a concern in many applications. Attribution-based explanations~\cite{1,2,3} offer a general understanding of model predictions by localizing objects within a broad context or emphasizing distinct regions crucial for classification. Yet, as noted in~\cite{4}, these explanations often fall short in conveying the specific patterns recognized by the model, identifying "where" patterns are but not "what" they are. Moreover, traditional attribution methods face limitations in fine-grained classification tasks, where subtle, localized features are critical to the model's predictions~\cite{5}. Consequently, regardless of whether the model's prediction is correct or incorrect, the highlighted regions often appear indistinguishable, thus diminishing their interpretative value for human users.
\begin{figure}[tb]
    \centering
    \includegraphics[width=0.9\linewidth]{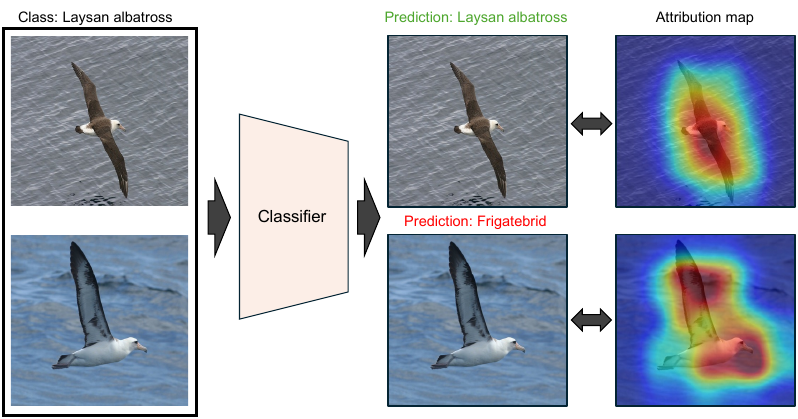}
    \caption{Traditional attribution-based explanation techniques often produce similar visual explanations for both correctly classified and misclassified samples.}
    \label{fig1}
\end{figure}

Fine-grained explanations are essential for understanding model predictions, especially when distinguishing between similar objects in real-world scenarios~\cite{6}. As the complexity of recognition tasks increases, the demand for more nuanced explanations becomes critical~\cite{7,8}. For instance, in Fig.~\ref{fig1}, while the highlighted pixels indicate the presence of the bird, it is difficult to ascertain which specific local features contribute to the model's correct prediction of "Laysan albatross" and which contribute to its incorrect prediction of "Frigatebird." Fine-grained features, such as the head, wings, or other distinctive parts, can reveal crucial differences between closely related classes, providing valuable insights into the model's decision-making process~\cite{9}. In simpler classification tasks, like distinguishing between cats and dogs, a coarse explanation may suffice; however, in complex cases, such as differentiating between bird species, coarse explanations lose their interpretive power as they often highlight the entire object without distinguishing between meaningful regions.

It is well-recognized that even experts can struggle to classify an object accurately when its distinguishing characteristics are subtle. However, identifying the correct class becomes significantly easier when multiple samples from other classes are available, allowing for comparative analysis of fine-grained features~\cite{10,11,12}. Contrastive explanations~\cite{13} enable this process by providing insights through comparisons between similar samples from different classes. In this work, we adopt a contrastive explanation approach specifically for misclassified samples. Particularly, our approach generates contrastive explanations using counterfactuals created from the misclassified samples themselves. By modifying the sample to alter the model's prediction to the correct class, we can identify the fine-grained features responsible for the initial misclassification.


A recent study~\cite{11} on contrastive counterfactual explanation (CCE) addresses the question, “Why P, rather than Q?” by exploring why a sample belongs to class P rather than class Q. This study visualizes attribution maps before and after applying a projected gradient descent (PGD) attack~\cite{14}. However, the generated counterfactuals in this method prompt a prediction change through imperceptible gradients, rendering the results theoretically unconvincing. Additionally, ~\cite{15} points out that even when PGD attacks succeed in altering predictions, the resulting attribution maps remain largely unchanged.

To our knowledge, this work is the first to emphasize the role of fine-grained features in model explanations. While fine-grained datasets have been utilized in previous counterfactual explanation research~\cite{16,17}, those methods do not specifically focus on fine-grained explanations. Notably, we adopt a non-generative approach similar to theirs for creating counterfactuals by leveraging existing data samples. However, our framework is explicitly designed to provide explanations at both the object-level and fine-grained feature level. 

In this paper, we propose a novel \textbf{F}ine-\textbf{G}rained \textbf{V}isual \textbf{C}ontrastive \textbf{E}xplanation (FG-VCE) framework designed to provide fine-grained contrastive explanations for misclassified samples using counterfactuals. This approach focuses on two key aspects: (1) identifying fine-grained features critical for prediction change, and (2) understanding the factors contributing to model misclassification. To achieve this, we calculate the contribution of each feature point through approximate Shapley values. Building on prior work~\cite{11}, we address the entanglement among feature maps by introducing a Saliency Partition module with spatially localized kernels. Additionally, our iterative counterfactual generation approach leverages a set of feature candidates that are highly contributive to each sample. In each iteration, we select the most contributive feature in the misclassified sample and match it to the candidate with the highest semantic similarity. This process repeats until the prediction changes, thus completing the counterfactual explanation. The proposed framework is non-generative, end-to-end, and adaptable across a wide range of fine-grained tasks and models. The primary contributions of this paper are as follows: 
\begin{itemize} 
    \item we present the first work on fine-grained CCE, proposing a framework that emphasizes individual feature contributions and benchmarks fine-grained performance on two widely-used datasets;
    \item our framework generates distinct, localized explanations without compromising semantic consistency, achieving efficiency and fidelity to the original sample through a non-generative, candidate-based approach; 
    \item the results highlight highly interpretable, fine-grained regions that provide clear insights for human users. Extensive ablation studies further demonstrate the effectiveness of the proposed framework.
\end{itemize}

\begin{figure*}[htb]
    \centering
    \includegraphics[width=\textwidth]{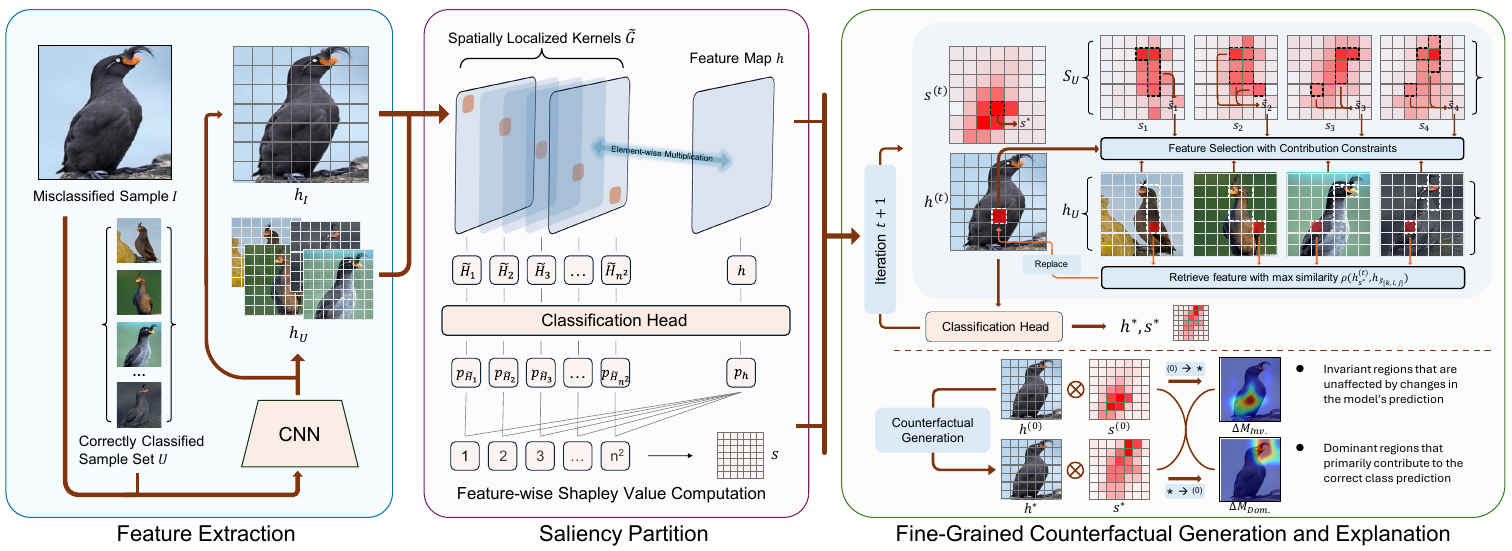}
    \caption{The proposed FG-VCE framework. The framework consists of three main stages: (1) feature extraction, in which the model processes both the misclassified sample and correctly classified reference samples to generate feature representations; (2) saliency partition, which computes the contribution of individual feature points using an approximation of the Shapley value; and (3) fine-grained contrastive counterfactual generation and explanation, where the most informative regions are selectively modified to produce a contrastive explanation that aligns with the target class while preserving semantic consistency.}
    \label{fig.framework}
\end{figure*}

\section{Related Works}

\subsection{Attributive Explanations}


Attributive explanations, a widely adopted approach in visual explanation techniques, aim to identify specific pixels or regions in an image that substantially influence a classifier's prediction. Saliency map-based methods~\cite{18,19,20,21,22} effectively highlight areas or features that contribute most to the model's decision-making. Earlier class activation-based frameworks focused primarily on gradient-based methods~\cite{1,23}, calculating the gradients of the classifier with respect to a particular input or network layer. However, the attribution maps produced by these methods are typically based on the last convolutional layer, often resulting in low-resolution and blurred target activation maps.

More recently, advanced attribution methods~\cite{5} have shown exceptional performance in fine-grained tasks by generating clear global attribution maps even within deep feature layers. Despite this progress, current global attribution explanations encounter challenges in explaining model classification decisions for classes with similar features, limiting their interpretability in fine-grained classification tasks.

\subsection{Visual Counterfactual Explanations}


The objective of visual counterfactual explanations is to identify the minimal feature region that, when altered, results in a change in classification decision. Adversarial attacks~\cite{24,25,26,27} illustrate this concept by incrementally adding noise to an image from class \( P \) until the classifier’s prediction shifts to class \( Q \). Recent advancements in generative methodologies~\cite{28,29,30}, especially those leveraging generative models, have shown considerable potential. These methods, often applied to datasets such as MNIST~\cite{31} and CelebA~\cite{32}, use Generative Adversarial Networks (GANs) to produce more interpretable explanations.

In non-generative counterfactual approaches, methods like~\cite{10} employ attribution maps to isolate critical regions, while approaches such as CoCoX~\cite{33} identify visual concepts whose addition or removal alters model predictions. Common strategies~\cite{16,17} use distractor images \( I' \) from class \( Q \) to identify regions in a query image \( I \) from class \( P \) to replace, shifting the model’s prediction to class \( Q \). This counterfactual generation technique is model-agnostic, requiring no additional training, making it versatile across classification models. However, exhaustive feature replacement can significantly reduce the efficiency of counterfactual generation.


\section{FG-VCE Framework}
To generate a counterfactual explanation, as illustrated in Fig.~\ref{fig.framework}, the proposed explanation framework comprises three components. Let $f$ represent the model, and $I$ denote the misclassified sample. Additionally, a correctly classified sample set $U$ is required. First, in feature extraction, both sample $I$ and set $U$ are input into the model, yielding feature representations $h$ and $h_U$, respectively. Then, the contribution of each feature point is computed through a Saliency Partition module, allowing refined Shapley value computation. Finally, a non-generative counterfactual explanation is generated by iteratively updating the most informative regions under similarity and semantic constraints, producing a detailed, fine-grained explanation. 


\subsection{Saliency Partition based Feature Contribution}
The traditional saliency map~\cite{1,2,23} highlights groups of local features but lacks the capacity to differentiate and quantify the individual importance of each feature to the model's prediction. Consequently, it is insufficient for providing a fine-grained explanation. To achieve a more granular interpretability, it is essential to avoid generating the saliency map as a holistic representation, and to minimize the interdependence between neighboring points in the feature space. Thus, we propose a Saliency Partition (SP) module to disentangle and emphasize the contributions of individual feature points.

Given a feature map $h \in \mathbb{R}^{n \times n}$, we aim to compute the contribution of each point. Inspired by the concept of the Shapley value~\cite{34}, rather than evaluating all possible subsets $S$ from the union of features, we approximate the contribution of a single point by subtracting the contribution without that point from the overall contribution, formulated as follows:
\begin{equation}
s = p_h \cdot \mathbf{1} - p_{\Tilde{H}},
\label{s}
\end{equation}
where $p_h \in \mathbb{R}$ is the model's prediction based on the entire feature map, representing the aggregate contribution, and $p_{\Tilde{H}}\in \mathbb{R}^{n\times n}$ is a contribution matrix, approximating the Shapley values for each feature point's impact on the prediction. Here, $\mathbf{1}$ is an all-ones matrix used to facilitate broadcasting across dimensions.

To compute \( p_{\Tilde{H}} \), a key challenge arises from the strong influence of neighboring features, which complicates isolating the contribution of individual feature points. Traditional Shapley formulas are insufficient for this purpose~\cite{35}. To disentangle neighboring features, we define coordinate grid matrices such that $X=[0~1~\dots~n-1]^T \mathbf{1}_{1\times n}$, and $Y=\mathbf{1}_{n\times 1}[0~1~\dots~n-1]$. We then introduce a set of Gaussian kernels $G \in \mathbb{R}^{n^2 \times n \times n}$. Let $\vec{c} \in \mathbb{R}^{n^2 \times 2}$ be a vector representing the center points for each Gaussian kernel, where each row of $\vec{c}$ specifies a unique center $(x_c,~y_c)$ ranging from 0 to $n-1$ in both dimensions. Using broadcasting, we construct Gaussian matrices $G_{[k,~:,~:]}$ for each center $\vec{c}_k = (x_{c_k},~y_{c_k})$ as follows:
\begin{equation}
G_{[k,~:,~:]} = \exp\left( -\frac{(X - x_{c_k})^2 + (Y - y_{c_k})^2}{2\sigma^2} \right),
\label{g}
\end{equation}
where each slice $G_{[k,~:,~:]}$ corresponds to a Gaussian distribution centered at $\vec{c}_k$.

Additionally, we define the reversed Gaussian kernel matrix $\Tilde{G} \in \mathbb{R}^{n^2 \times n \times n}$ by taking $\mathbf{1} - G$:
\begin{equation}
\Tilde{G} = \mathbf{1} - G,
\label{k}
\end{equation}
where $\Tilde{G}_{[:,~i,~j]} = \mathbf{1} - G_{[:,~i,~j]}$, and each $\Tilde{G}_{[:,~i,~j]} \in \mathbb{R}^{n \times n}$ represents a matrix in which the region around position $(i,~j)$ is attenuated. We refer to $\Tilde{G}$ as a set of spatially localized kernels, which is composed of slices $\Tilde{G}_{[k,~:,~:]}$, each of which has values close to 1, except for a "dip" around the corresponding center $\vec{c}_k$.

The tensor $\Tilde{H} \in \mathbb{R}^{n^2 \times n \times n}$ is obtained by performing an element-wise multiplication of $h$ with each $\Tilde{G}_{[k,~:,~:]}$:
\begin{equation}
\Tilde{H}[k,~:,~:] = h \odot \Tilde{G}_{[k,~:,~:]},
\label{h}
\end{equation}
where each $\Tilde{H}[k,~:,~:]$ has its feature expression suppressed around $\vec{c}_k$, creating a “partition” that emphasizes the feature of interest. By adjusting the scale factor $\sigma$, we can control the extent and smoothness of this partitioning effect.

Subsequently, each partitioned feature map $\Tilde{H}[k,~:,~:]$ is passed through the model to obtain a prediction. Reshaping it to match the size of the feature map yields the contribution matrix of predictions $p_{\Tilde{H}}$, where each entry corresponds to a prediction for a specific partitioned feature map. By substituting Eq.~\ref{h} into Eq.~\ref{s}, we obtain the refined Shapley value $s$, where each element represents the contribution of a specific region in the feature map.

\subsection{Fine-Grained Counterfactual Generation}
Our approach for non-generative counterfactual generation is conceptually straightforward. Inspired by~\cite{8,9}, we aim to increase the probability of the target class by replacing specific features of the misclassified sample with those from correctly classified instances. We refer to the features targeted for replacement as \textbf{target features} and the substituting features as \textbf{candidate features}. However, prior to implementation, two primary challenges must be addressed. The first challenge involves determining which features should be replaced, and the second requires identifying suitable replacement features that preserve semantic and perceptual consistency.

To address the first challenge, we intuitively select the feature with the highest Shapley value for replacement, as this feature likely plays a critical role in the misclassification. For the second challenge, we consider all features in $h_U$ as candidates and select the one with the highest semantic similarity to the target feature, ensuring it also contributes to its correct class prediction. Once the target and candidate features are identified, we replace the target feature with the candidate feature, denoted as $h^*$, and input the modified feature set into the model to obtain a new prediction.

A single replacement, however, may not be sufficient to change an incorrect prediction to a correct one. Consequently, we apply this replacement procedure iteratively until the prediction aligns with the correct class. Let $t$ denote the number of iterations required for the prediction change; the resulting modified feature is then represented as $h^{(t)}$, with $h^*$ indicating the final feature responsible for driving the prediction change.


In iteration \( t \), let the feature with the highest Shapley value be denoted as \( h_{s^*}^{(t)} \), where \( s^* = \max s^{(t)} \). To achieve fine-grained explainability, we aim to identify a feature that retains perceptual and semantic similarity to the original feature \( \max s^{(t)} \). Ideally, we would compute the similarity between all features in the set \( U \) and select the one with the highest similarity. However, this approach is computationally intensive and fails to consider the contribution of these features to their respective predictions. 

Thus, let \( S_U = \{s_1,~s_2,~s_3,~\dots\} \) represent the Shapley values for the set \( U \). We retain only the top-\( m \) values for each element to construct a refined set \( \Tilde{S}_U \), such that \( \Tilde{S}_U = \{\Tilde{s}_k~|~\Tilde{s}_k = \text{Top}_m(s_k),~\forall s_k \in S_U\} \). This approach allows us to concentrate on features with substantial contributions to model predictions. Notably, while \( s^{(t)} \) for \( h^{(t)} \) must be computed in each iteration, \( \Tilde{S}_U \) is calculated only once in the initial iteration and can be reused in all subsequent iterations.

Thus, the objective function can be formulated as:
\begin{equation}
\mathcal{L}_{\text{sim}} = \rho(h_{s^*}^{(t)},~h_{\Tilde{s}_{[k,~i,~j]}}),
\label{l_sim}
\end{equation}
where $\rho$ represents cosine similarity, and $h_{s^*}^{(t)}$ and $h_{\Tilde{s}_{[k,~i,~j]}}$ are target feature and candidate feature, respectively. The subscript $U$ is omitted for candidate feature for simplicity, but shall not raise confusion based on contexts. Furthermore, we want the replacement to yield a substantial improvement in the model's prediction for the correct class. The classification loss is defined as:
\begin{equation}
\mathcal{L}_{\text{cls}} = \sum_c y_c\log f_c(h^{(t)}~|~t,s^{*},\Tilde{s}_{[k,~i,~j]}),
\label{l_cls}
\end{equation}
where $f_c$ is the classifier that provides the prediction for the feature maps, and $y_c$ is the label for class $c$, and is defined as:
\begin{equation}
        y_c =\begin{cases} 1, & c = a \\
              0, &  c \neq a
     \end{cases},
\end{equation}
where $a$ is the correct class of $h$. Our goal is to maximize the joint objective of Eq.~\ref{l_sim} and Eq.~\ref{l_cls}, which is expressed as:
\begin{equation}
\mathcal{L}_{\text{tot}} = \mathcal{L}_{\text{sim}} + \mathcal{L}_{\text{cls}},
\end{equation}
such that the optimal indices are given by:
\begin{equation}
[k,~i,~j]^{(t)}= \arg \max \mathcal{L}_{\text{tot}}.
\label{argmax}
\end{equation}

\subsection{Contrastive Explanation for Misclassification}

Inspired by~\cite{11}, we generate visual explanations for model misclassifications by examining the differences in feature representations before and after counterfactual generation.

First, to identify the \textbf{invariant features}, denoted as \( \Delta M_{Inv.} \), which contribute to the incorrect class consistently even when the model transitions from an incorrect to a correct prediction, we subtract the weighted feature map after counterfactual generation from the original feature map:
\begin{align}
\Delta M_{Inv.} = s^{(0)} \cdot \frac{\sigma_N(h^{(0)} - h^{*})}{\sum_{i,j} \sigma_N(h^{(0)}_{i,j} - h^{*}_{i,j})},
\label{eq6}
\end{align}
where \( \sigma_N \) is the ReLU activation function. These invariant regions serve as significant contributors to the incorrect class decision, indicating that the model’s misclassification is due to this particular region.

Secondly, by reversing the order of subtraction (i.e., subtracting the activation map prior to the counterfactual generation from the one after), we identify regions that contribute predominantly to the correct class. These regions, emphasized following correction of the misclassification, represent the \textbf{dominant features} of the correct class decision. This can be defined as \( \Delta M_{Dom.} \), formulated as:
\begin{align}
\Delta M_{Dom.} = s^{*} \cdot \frac{\sigma_N(h^{*} - h^{(0)})}{\sum_{i,j} \sigma_N(h^{*}_{i,j} - h^{(0)}_{i,j})},
\label{eq7}
\end{align}
where these regions highlight the key features the model leverages to correct its prediction to the correct class after an initial misclassification.

\begin{table*}[ht]
    \centering
    \caption{Performance comparison of various methods utilizing ResNet-50 and VGG-19 backbones. The evaluation metrics include Insertions (Ins.), Deletions (Del.), and the Compact Activation Score (\( \xi \)). For Ins. and \( \xi \), higher values indicate better performance, while for Del., lower values are preferred.}
    \resizebox{0.9\textwidth}{!}{
        \begin{tabular}{lrrr|rrr|rrr|rrr}
        \toprule
            & \multicolumn{6}{c}{CUB-200-2011} & \multicolumn{6}{c}{Stanford Dogs} \\
            \cmidrule(lr){2-7} \cmidrule(lr){8-13}
            & \multicolumn{3}{c}{ResNet-50} & \multicolumn{3}{c}{VGG-16} & \multicolumn{3}{c}{ResNet-50} & \multicolumn{3}{c}{VGG-16} \\
            \cmidrule(lr){2-4} \cmidrule(lr){5-7} \cmidrule(lr){8-10} \cmidrule(lr){11-13}
            Method & Ins. $\uparrow$ & Del. $\downarrow$ & $\xi\uparrow$~ & Ins. $\uparrow$ & Del. $\downarrow$ & $\xi\uparrow$~ &  Ins. $\uparrow$ & Del. $\downarrow$ &  $\xi\uparrow$~ & Ins. $\uparrow$ & Del. $\downarrow$ &  $\xi\uparrow$~ \\ 
            \midrule
            GBP~\cite{39} & 39.71 & 14.72 & 5.76 & 30.12 & 11.07 & 4.85 & 37.02 & 12.10 & 6.12 & 35.12 & \textbf{11.13} & 5.41 \\
            IG~\cite{38} & 35.38 & 13.56 & 4.21 & 26.58 & 14.46 & 4.67 & 33.67 & 12.71 & 5.32 & 31.44 & 13.69 & 5.14 \\
            RISE~\cite{40} & 35.49 & 17.27 & 6.02 & 29.05 & 15.64 & 6.02 & 32.61 & 15.11 & 5.83 & 32.58 & 14.32 & 5.66 \\
            EP~\cite{41} & 32.16 & 14.87 & 5.92 & 27.05 & 13.35 & 5.78 & 30.03 & 12.55 & 5.93 & 29.24 & 12.63 & 5.86 \\
            GradCAM~\cite{1} & 36.43 & 18.61 & 4.58 & 24.87 & 11.95 & 4.92 & 35.13 & 16.32 & 5.11 & 33.25 & 15.12 & 5.01 \\
            GradCAM++~\cite{23} & 36.58 & 18.51 & 4.50 & 25.01 & 11.46 & 5.06 & 36.24 & 17.51 & 5.46 & 35.63 & 16.44 & 5.54 \\
            ScoreCAM~\cite{2} & 37.09 & 18.56 & 3.95 & 27.46 & 11.94 & 5.49 & 37.43 & 17.22 & 5.75 & 30.85 & 12.73 & 5.33 \\
            FG-CAM~\cite{5} & \underline{41.31} & 14.55 & 7.21 & 39.26 & 14.21 & 6.45 & \underline{39.62} & 13.07 & 7.01 & 38.53 & 12.88 & 6.95 \\
            CALM~\cite{42} & 34.18 & 14.23 & 5.01 & 31.82 & 11.91 & 5.41 & 32.93 & 13.27 & 5.10 & 30.26 & 12.35 & 5.03 \\
            \midrule
            CE~\cite{43} & 37.18 & 17.92 & 4.59 & 28.39 & 12.57 & 5.12 & 36.12 & 16.84 & 5.53 & 35.64 & 15.78 & 5.89 \\
            SCOUT~\cite{10} & 33.78 & 13.01 & 5.88 & 31.26 & 13.12 & 5.13 & 32.19 & \textbf{11.24} & 6.19 & 31.03 & 13.49 & 5.65 \\
            CCE~\cite{11} & 40.47 & \underline{11.14} & \underline{7.85} & \underline{32.86} & \underline{10.79} & \underline{6.32} & 39.57 & 12.54 & \underline{9.12} & \underline{39.02} & 12.31 & \underline{7.56} \\
            \midrule
            \rowcolor[gray]{0.9}
            \textbf{FG-VCE (Ours)} & \textbf{43.02} & \textbf{10.27} & \textbf{9.53} & \textbf{34.25} & \textbf{9.89} & \textbf{8.96} & \textbf{42.15} & \underline{12.08} & \textbf{10.45} & \textbf{41.58} & \underline{12.06} & \textbf{8.32} \\
            \bottomrule
        \end{tabular}
    }
    \label{tab:performance_comparison}
\end{table*}

\begin{figure*}[htb]
    \centering
 \includegraphics[width=1.0\textwidth]{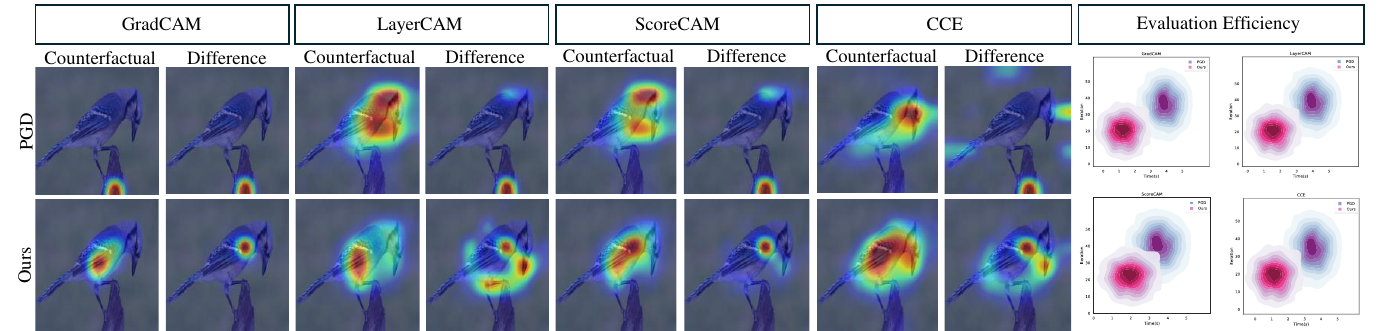}
    \caption{Qualitative comparison of fine-grained attribution maps generated by PGD and our proposed method. The efficiency distribution of misclassified samples under both PGD and our method is presented on the right.}
    \label{fig2}
\end{figure*}
\section{Experiments}
\subsection{Setups}
\textbf{Datasets.}  
The experimental evaluation was conducted on the CUB-200-2011~\cite{36} and Stanford Dogs~\cite{37} datasets. CUB-200-2011 is a richly annotated, fine-grained dataset focused on bird species, containing 200 unique bird categories. Each bird image has attributes assigned to specific parts, resulting in a dataset of 11,988 images, split into 5,994 images for training and 5,794 images for validation. The Stanford Dogs dataset comprises dog images annotated with keypoint locations for 24 parts, totaling 20,580 images. This dataset is divided into 12,000 images for training and 8,580 images for validation.


\textbf{Baseline Comparisons.} To assess the effectiveness of the FG-VCE method, we compared it against a range of state-of-the-art attribution map generation approaches. These include propagation-based methods (Integrated Gradients (IG)~\cite{38} and Guided Backpropagation (GBP)~\cite{39}), perturbation-based methods (RISE~\cite{40} and EP~\cite{41}), activation-based methods (GradCAM~\cite{1}, GradCAM++~\cite{23}, ScoreCAM~\cite{2}, FG-CAM~\cite{5}, and CALM~\cite{42}), and contrastive counterfactual-based methods (CCE~\cite{11}, CE~\cite{43}, and SCOUT~\cite{10}). Additionally, while we adopt the latest PGD attack-based method~\cite{11} as a baseline for visual comparison, our approach does not rely on adversarial training. In our experiments, we primarily demonstrate the differences between these two types of visual explanation methods and highlight the advantages of our approach.


\textbf{Implementation Details.} To evaluate the generalizability of our approach, we employed ResNet-50~\cite{44} and VGG-16~\cite{45} as backbone architectures. Feature replacement was performed on the output of the last convolutional layer for both models. We set the maximum number of iterations \( t \) to 100, exceeding this limit indicated failure in counterfactual generation. The number of samples in the correctly classified set \( U \) was set to 20, balancing performance and computational cost. Similarly, the scale factor \( \sigma \) was fixed at 0.8. All experiments were conducted on an NVIDIA RTX A5000 GPU with 24 GB memory. The implementation is available at \href{https://github.com/LIMTONG/FG-VCE}{https://github.com/LIMTONG/FG-VCE}.

\subsection{Evaluation Metrics}
\textbf{Insertion and Deletion Tests.}  
We conducted insertion and deletion tests as described in~\cite{11,46} on 500 images from standard fine-grained task datasets to evaluate the performance of various saliency methods. The deletion metric is based on the assumption that removing pixels or regions most relevant to a target class should significantly reduce classification scores. In this test, we progressively replace \(1.8\%\) of the original image’s pixels with a highly blurred version, following the order defined by the saliency map, until all pixels are removed. Conversely, the insertion metric gradually reintroduces the original image content into a blurred image. This approach generates images that better adhere to the data manifold, reducing adversarial effects. In this test, we replace \(1.8\%\) of the blurred image’s pixels with the original pixels incrementally until the image is fully restored.

\textbf{Compact Activation Score ($\xi$).}  
Previous evaluation metrics for saliency maps have primarily focused on global features, limiting their effectiveness in validating local features with fine-grained precision. To address this gap, we propose a novel metric tailored specifically for evaluating the fine-grained quality of saliency maps, termed the Compact Activation Score (\(\xi\)):
\begin{align}
\xi = C \times \frac{P_T}{P_A},
\label{eq8}
\end{align}
where \( C \) represents the ratio between the sum of fine-grained class activation features, denoted as \(\Delta M_{\text{Dom.}}\), and the normalized sum of global class activation features, \(\Delta h^{*}\). Here, \( P_T \) and \( P_A \) denote the total number of elements in the global feature map and the count of non-zero elements within the fine-grained saliency map \(\Delta M_{\text{Dom.}}\), respectively. This formulation implies that higher scores are achieved when the fine-grained saliency map occupies a more compact area yet significantly influences the class prediction, thereby emphasizing the metric’s capability to assess localization precision and relevance of the identified salient regions.

\begin{figure*}[htb]
    \centering
    \includegraphics[width=\textwidth]{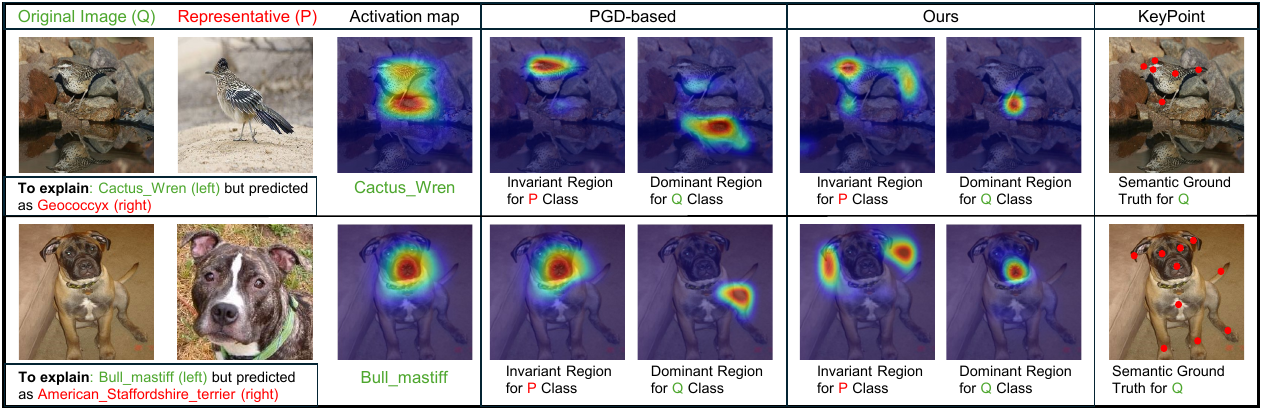}
    \caption{Evaluating the rationality of visual contrastive explanations. Answer the question 'Why P (misclassification class) rather than Q (correct class)?' based on invariant regions and dominant regions in fine-grained classification task.}
    \label{cce}
\end{figure*}




\subsection{Analysis}
\subsubsection{Quantitative Comparison}
To validate the significance of the fine-grained saliency maps $\Delta M$ generated by our method for class decisions, we compared its performance against state-of-the-art methods, as presented in Table~\ref{tab:performance_comparison}. The results demonstrate that, for fine-grained tasks, our method provides superior explanations, outperforming most competing approaches in terms of deletion and insertion metrics. Notably, on the CUB-200-2011 dataset, our method achieved the best deletion and insertion values of \textbf{43.02\%} and \textbf{10.27\%}, respectively. Furthermore, the compact activation score reached a peak value of \textbf{9.53}, indicating that the fine-grained explanations generated by our method make the most significant contribution to class decisions.

In terms of visualization, as depicted in Fig.~\ref{fig2}, our method consistently produces more coherent explanations compared to PGD-based approaches across different attribution functions. While minor variations in performance are observed depending on the specific attribution function, the explanations generated by our method predominantly focus on the target object, resulting in more interpretable and reasonable visual explanations. Finally, we also evaluated the computational complexity associated with generating explanations across multiple methods. As shown in Table~\ref{complexity}, our approach demonstrates certain advantages in terms of both time consumption and GPU memory usage during the explanation generation process.

\begin{table}[h]
\centering
\caption{Evaluation of the computational complexity with a single NVIDIA RTX A5000 GPU 24GB.}
\label{complexity}
    \resizebox{\linewidth}{!}{
    \large
    \begin{tabular}{*{10}{c}}
        \toprule
        Complexity /\ Methods & CE~\cite{43} & CVE~\cite{16}  & CCE~\cite{11} & RISE~\cite{40} & \textbf{Ours} \\
        \midrule
        Time usage (s) & 8.8 & 6.7 & 13.5 &16.9 & \textbf{5.2} &   \\
        Memory usage (GB) & 7.6 &6.4& 5.4 & 10.6 & \textbf{3.2} &  \\
        \bottomrule
    \end{tabular}
    }
    \vspace{-1em}
\end{table}





\subsubsection{Visual Contrastive Explanation}

We conducted evaluation experiments to compare our method with prior works in visual comparative explanations.  As illustrated in Fig.~\ref{cce}, the activation maps generated before and after feature replacements effectively address the following interpretability question: for instance, in the first column, why did the model incorrectly classify the image as \textbf{Geococcyx (P)} instead of \textbf{Cactus Wren (Q)}? 

Analyzing the generated invariant regions reveals that when the prediction shifts from class \( P \) to class \( Q \), only the head features are crucial for class \( P \), explaining why the model initially favored class \( P \). Conversely, when the prediction changes to class \( Q \), the generated dominant region highlights the foot area, which is the key determinant for class \( Q \). This effectively explains why the image was not initially classified as class \( Q \).

Compared to previous PGD-based visual comparative explanation methods, the fine-grained saliency maps produced by our approach integrate key point information of the relevant features. This enhancement allows our method to provide a more comprehensive and precise explanation of model decisions, particularly in fine-grained visual tasks.

\begin{table}[t] 
    \centering
    \caption{The impact of SP module on attribution map-based contrastive explanation methods.}
    \resizebox{\columnwidth}{!}{ 
    \begin{tabular}{lrrr|rrr}
        \toprule
        \multirow{2}{*}{Method} & \multicolumn{3}{c}{ResNet-50} & \multicolumn{3}{c}{VGG-19} \\
        \cmidrule(lr){2-4} \cmidrule(lr){5-7}
         & Ins. $\uparrow$ & Del. $\downarrow$ & $\xi\uparrow$~ & Ins. $\uparrow$ & Del. $\downarrow$ & $\xi\uparrow$~ \\
        \midrule
       
        CE~\cite{43} & 37.18 & 17.92 & 4.59 & 28.39 & 12.57 & 5.12 \\
        CE~\cite{43} \textbf{+ SP} & \textbf{38.43} & \textbf{16.29}&  \textbf{6.21}&\textbf{ 30.04} & \textbf{5.89} &\textbf{11.71} \\
        SCOUT~\cite{10} & 33.78 & 13.01 & 5.88 & 31.26 & 13.12 & 5.13 \\
        SCOUT~\cite{10} \textbf{+ SP} &\textbf{34.12}  &  \textbf{11.96}& \textbf{7.14}& \textbf{31.36} &\textbf{13.97}  & \textbf{6.87}\\
        CCE~\cite{11} & 40.47 & 11.14 & 7.85 & 32.86 & 10.79 & 6.32 \\
        CCE~\cite{11} \textbf{+ SP}& \textbf{41.39} & \textbf{10.52} &\textbf{9.16}  & \textbf{34.16} & \textbf{10.21} &  \textbf{8.75}\\
        \midrule 
        \textbf{FG-VCE} w/o \textbf{SP} & 41.71 & 10.14 & 8.31 & 33.81 & 10.93 & 7.20 \\
        \rowcolor[gray]{0.9}
        \textbf{FG-VCE (Ours)} & \textbf{43.02} & \textbf{9.27} & \textbf{9.53} & \textbf{34.25} & \textbf{9.89} & \textbf{8.96}\\
        \bottomrule
    \end{tabular}
    }
    \label{ag}
\end{table}

\begin{figure}[t]
    \centering
    \includegraphics[width=\linewidth]{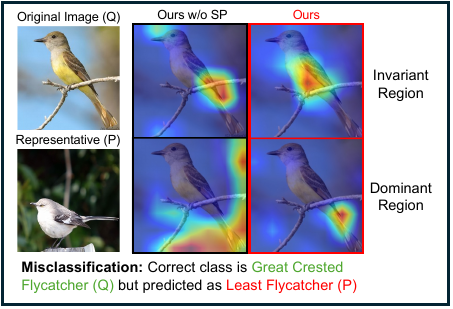}
    \caption{The impact of refining shapley operations on fine-grained saliency maps (invariant and dominant region) when generating explanations for misclassifications.}
    \label{fig.exp3}
\end{figure}
\subsubsection{Ablation Study}
In this study, we conducted ablation research through two main experiments. First, we evaluated the impact of proposed SP module on the quality of generated explanations. Second, we analyzed how the presence or absence of the SP module, along with the application of Top-$m$ selection based on Shapley values, affects the efficiency of generating explanations during feature unit replacement process.
\begin{figure}[t]
    \centering
    \includegraphics[width=\linewidth]{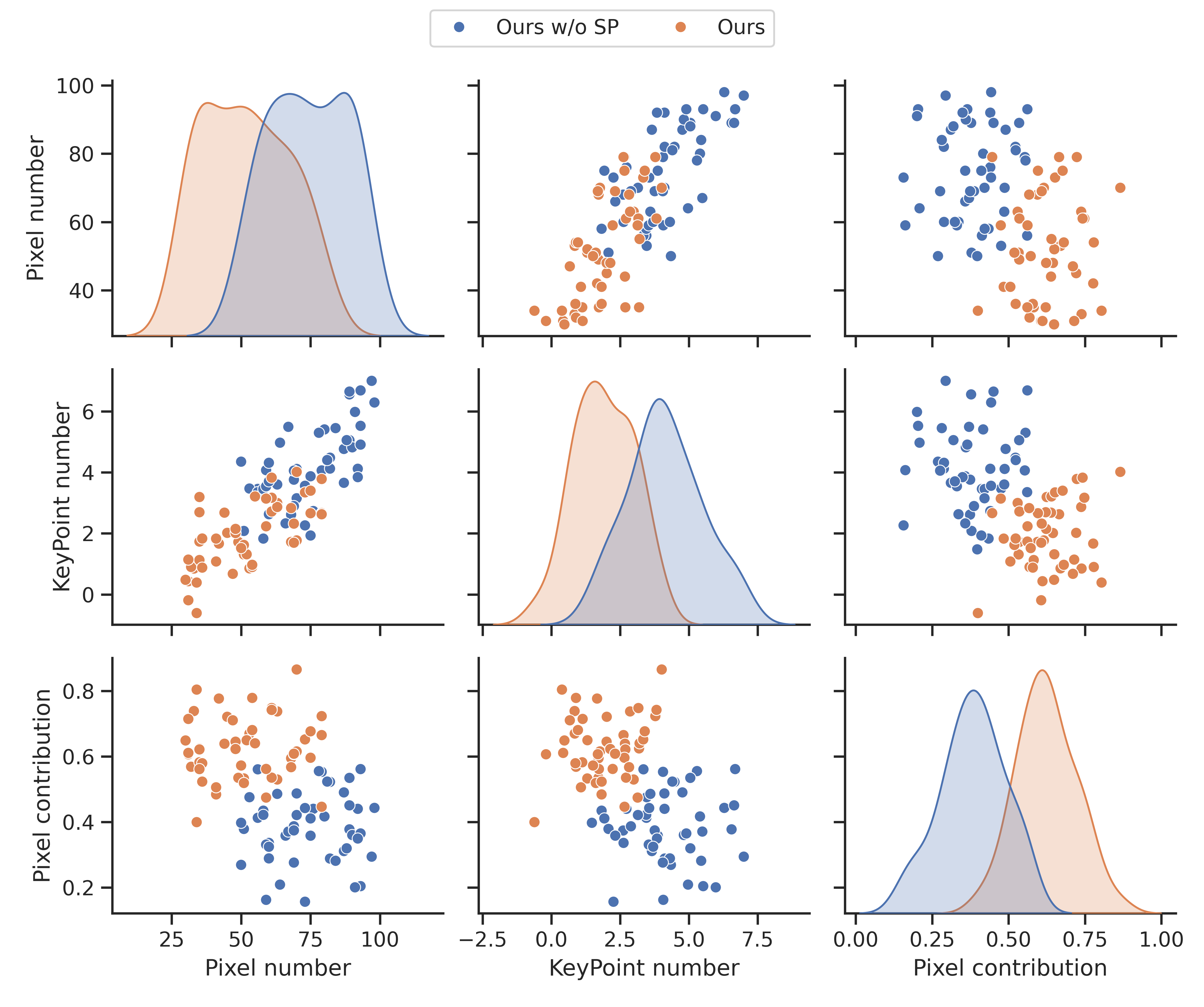}
    \caption{Evaluation of fine-grained effects of shapley value refinement on fine-grained saliency maps.}
    \label{fig.exp1}
\end{figure}

To verify the effect of the SP process on contrastive explanation generation, we integrated this process into several contrastive explanation methods and observed the quality of the resulting fine-grained explanations. As shown in Table~\ref{ag}, the inclusion of the SP module consistently improved the quality of fine-grained explanations. To facilitate understanding of these differences, Fig.~\ref{fig.exp3} demonstrates that the addition of the SP module results in more accurate and fine-grained semantic localization of the Invariant and Dominant regions, making the explanations of misclassifications more comprehensible. Furthermore, we conducted a detailed evaluation of the impact of this process on fine-grained explanations. As shown in Fig.~\ref{fig.exp1}, we performed fine-grained assessments on 50 fine-grained explanations generated under conditions with and without SP module. These evaluations considered the number of pixels, the number of keypoints included, and the pixel contribution. The results indicate that, with the SP module, the generated fine-grained explanations contain fewer total attribution pixels and fewer keypoints, while the average class contribution per pixel is higher. Naturally, extreme cases where no Keypoints are included may occur, and we exclude such explanations from the evaluation. This suggests that the scope of fine-grained explanations is reduced, focusing on a smaller number of local features, where each pixel within the region is more important for class decisions.

\begin{figure}[tb]
    \centering
    \begin{minipage}{\linewidth}
        \centering
        \begin{subfigure}[t]{0.485\textwidth} 
            \centering
            \includegraphics[width=\textwidth]{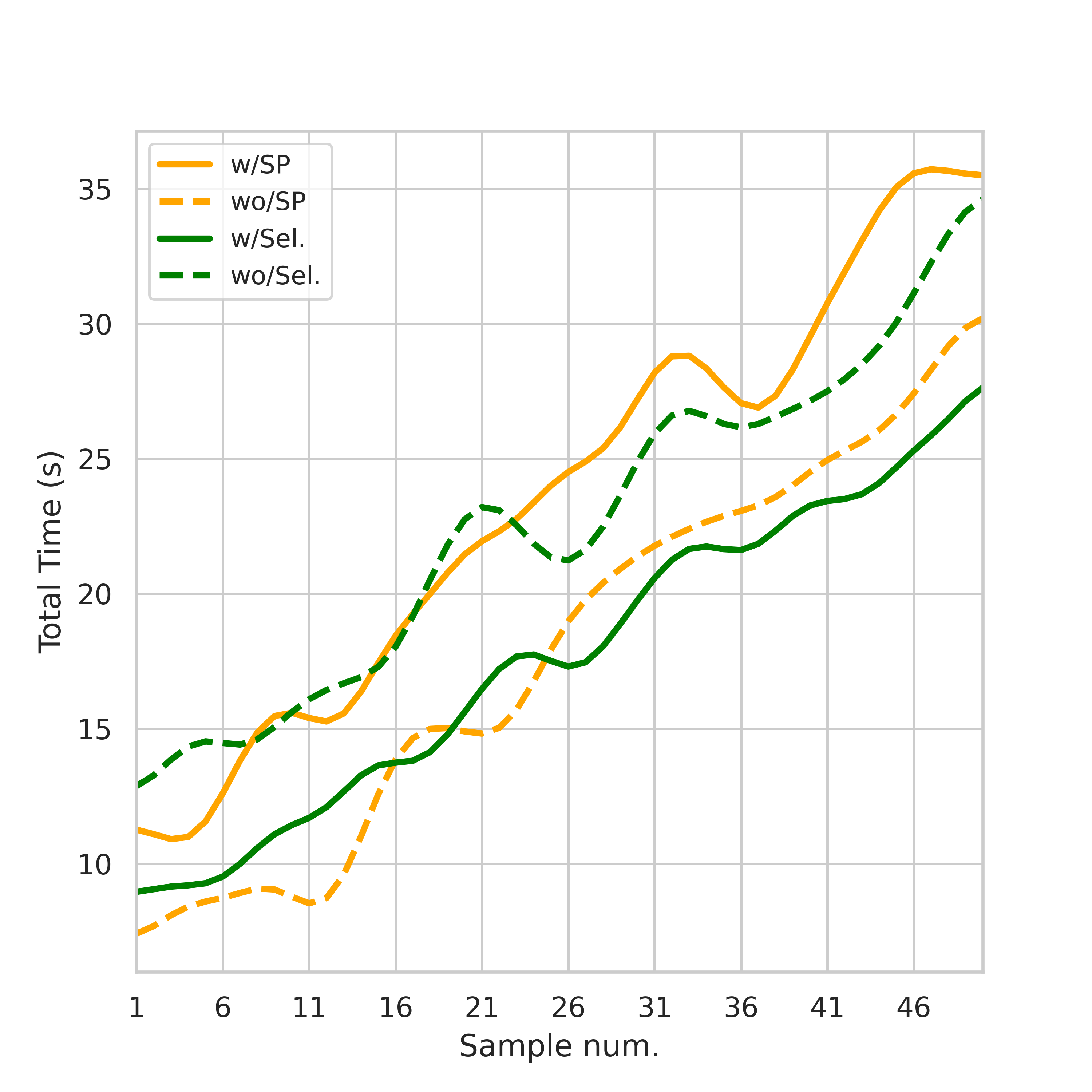}
            \caption{}
            \label{line}
        \end{subfigure}
        \hspace{0.01\textwidth} 
        \begin{subfigure}[t]{0.485\textwidth} 
            \centering
            \includegraphics[width=\textwidth]{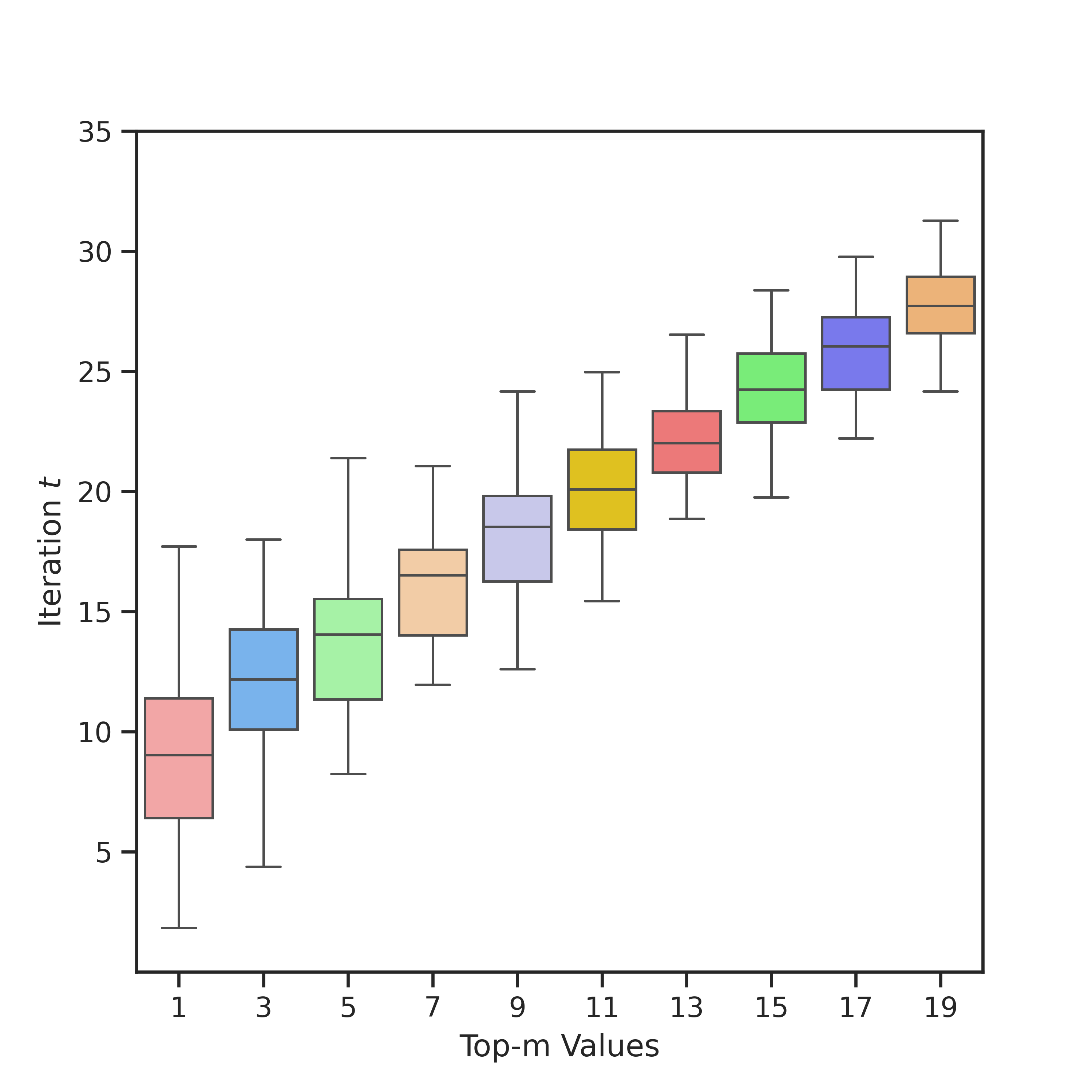}
            \caption{}
            \label{box}
        \end{subfigure}
        \caption{The impact of SP and Top-$m$ feature selection (Sel.) on explanation time and number of iteration.}
        \label{fig.combined}
    \end{minipage}
\end{figure}

Secondly, to evaluate the impact of the SP  module and Top-$m$ feature selection on the efficiency of explanation generation, we randomly selected 50 misclassified samples and recorded the total explanation generation time under the presence or absence of these conditions. As shown in Fig.~\ref{fig.combined} (a), the SP module increases the overall explanation generation time, while the application of Top-$m$ feature selection improves efficiency. Additionally, we investigated the relationship between the value of Top-$m$ and the number of iterations required to replace feature units in misclassified samples until a class change occurs. As shown in Fig.~\ref{fig.combined} (b), with an increase in the Top-$m$ value, the required iteration count tends to cluster in higher ranges.

\section{Conclusion}
We propose FG-VCE as a framework to provide visual explanations for misclassifications in fine-grained classification models. By generating distinctive local saliency maps, our approach addresses the critical question: "Why was the sample misclassified as class \( P \) rather than the correct class \( Q \)?" Through an iterative counterfactual generation process, FG-VCE produces contrastive saliency maps while preserving semantic integrity and contribution to class prediction. 

To further enhance interpretability in fine-grained classification tasks, we introduce a Saliency Partition (SP) module designed to disentangle the independent contributions of individual feature points. Experimental results demonstrate that our method provides more fine-grained and human-readable explanations for different predictions, enabling a deeper understanding of model behavior. In future work, we focus on the adaptability of this explanation method within transformer architectures.
\newpage
\section*{Acknowledgement}
This work was supported by the Institute of Information \& Communications Technology Planning \& Evaluation (IITP) grant, funded by the Korea government (MSIT) (No. RS-2019-II190079, Artificial Intelligence Graduate School Program (Korea University), No. RS-2024-00457882, AI Research Hub Project, and No. RS-2022-II220984, Development of Artificial Intelligence Technology for Personalized Plug-and-Play Explanation and Verification of Explanation).


\begin{thebibliography}{99}


\bibitem{1}R. Selvaraju et al., "Grad-cam: Visual explanations from deep networks via gradient-based localization,"in \textit{Proc. IEEE Int. Conf. Comput. Vis. (ICCV)} pp. 618-626, 2017.

\bibitem{2}H. Wang et al., “Score-CAM: Score-weighted visual explanations for convolutional neural networks,” in \textit{Proc. IEEE Conf. Comput. Vis. Pattern Recognit.(CVPR)}, pp. 24–25, 2020.

\bibitem{3}P. T. Jiang et al., "Layercam: Exploring hierarchical class activation maps for localization," \textit{IEEE Trans. Image Process.}, vol. 30, pp. 5875-5888, Jun. 2021.

\bibitem{4} R. Achtibat \textit{et al.}, "From attribution maps to human-understandable explanations through concept relevance propagation," \textit{Nat. Mach. Intell.}, vol. 5, no. 9, pp. 1006-1019, 2023.


\bibitem{5}C. Qiu, F. Jin, and Y. Zhang, "Empowering CAM-Based Methods with Capability to Generate Fine-Grained and High-Faithfulness Explanations, in \textit{Proc. AAAI Conf. Artif. Intell. (AAAI)}, pp. 4587-4595, 2024.


\bibitem{6} Y. Lyu et al., "Dime: Fine-grained interpretations of multimodal models via disentangled local explanations," in \textit{Proc. AAAI/ACM Conf. AI, Ethics, Soc.}, pp. 455-467, 2022.

\bibitem{7} R. Du \textit{et al.}, "Multi-View Active Fine-Grained Visual Recognition," in \textit{Proc. IEEE/CVF Int. Conf. Comput. Vis. (ICCV)}, pp. 1568-1578, 2023.


\bibitem{8} S. Huang, X. Wang, and D. Tao, "Stochastic partial swap: Enhanced model generalization and interpretability for fine-grained recognition," in \textit{Proc. IEEE/CVF Int. Conf. Comput. Vis. (ICCV)}, pp. 620-629, 2021.

\bibitem{9} Y. Chen, Y. Bai, W. Zhang, and T. Mei, "Destruction and Construction Learning for Fine-Grained Image Recognition," in \textit{Proc. IEEE/CVF Conf. Comput. Vis. Pattern Recognit. (CVPR)}, pp. 5157-5166, 2019.




\bibitem{10}P. Wang and V. Nuno, "Scout: Self-aware discriminant counterfactual explanations," in \textit{Proc.
IEEE/CVF Conf. Comput. Vis. Pattern Recognit. (CVPR)}, pp. 8981-8990, 2020.


\bibitem{11}X. Wang et al., "Counterfactual-based Saliency Map: Towards Visual Contrastive Explanations for Neural Networks," in \textit{Proc. IEEE Int. Conf. Comput. Vis. (ICCV)}, pp. 2042-2051, 2023.


\bibitem{12} W. Xie \textit{et al.}, "Two-stage holistic and contrastive explanation of image classification," in \textit{Proc. Uncertainty Artif. Intell. (UAI)}, pp. 2335-2345, 2023.

\bibitem{13}P. Wang and V. Nuno, "A generalized explanation framework for visualization of deep learning model predictions," \textit{IEEE Trans. Pattern Anal. Mach. Intell.}, vol. 45, no. 8, pp. 9265-9283, Aug. 2023.


\bibitem{14} A. Madry et al., "Towards deep learning models resistant to adversarial attacks," 2017, \textit{ arXiv:1706.06083}.

\bibitem{15}J. T. Springenberg et al.," Striving for Simplicity: The All Convolutional Net," in \textit{Proc. Int. Conf. Learn. Represent. workshop Track(ICLR Workshop Track)}, 2015.

\bibitem{16}Y. Goyal et al., "Counterfactual visual explanations," in \textit{Proc. Int. Conf. Mach. Learn. (ICML)}, pp. 2376-2384, 2019.

\bibitem{17}S. Vandenhende et al., "Making heads or tails: Towards semantically consistent visual counterfactuals," in \textit{Proc. Eur. Conf. Comput. Vis. (ECCV)},  pp. 261-279, 2022.






\bibitem{18}S. Srinivas and F. Fleuret, “Full-gradient representation for neural network visualization,” in \textit{ Proc. Int. Conf. Neural Inf. Process. Syst. (NeurIPS)}, pp. 4126–4135, 2019.

\bibitem{19}R. Shi, T. Li, and Y. Yasushi, "Group visualization of class-discriminative features," \textit{Neural Netw.}, vol. 129, pp. 75-90, May. 2020.

\bibitem{20}R. Chen et al., "Less is more: Fewer interpretable region via submodular subset selection," 2024, \textit{ arXiv:2402.09164}. 

\bibitem{21}W. J. Nam, J. Choi, and S-W. Lee, "Interpreting deep neural networks with relative sectional propagation by analyzing comparative gradients and hostile activations," in \textit{Proc. AAAI Conf. Artif. Intell. (AAAI)}, pp. 11604-11612,  2021.

\bibitem{22} W. J. Nam, S. Gur, J. Choi, L. Wolf, and S-W. Lee,  "Relative attributing propagation: Interpreting the comparative contributions of individual units in deep neural networks," in \textit{Proc. AAAI Conf. Artif. Intell. (AAAI)}, pp. 2501-2508, 2020.

\bibitem{23}A. Chattopadhay et al., "Grad-cam++: Generalized gradient-based visual explanations for deep convolutional networks," in \textit{Proc. IEEE Winter Conf. Appl. Comput. Vis. (WACV)}, pp. 839--847, 2018.








\bibitem{24}J. Ukita and O. Kenichi, "Adversarial attacks and defenses using feature-space stochasticity,"  \textit{Neural Netw.}, Vol. 167, pp. 875-889, Aug. 2023.

\bibitem{25} S. Gulshad, S. Sadaf, and A. Smeulders, "Counterfactual attribute-based visual explanations for classification," \textit{Int. J. Multimed. Inf. Retr.}, vol. 10, pp. 127-140, 2021.

\bibitem{26}M. Hashemi and F. Ali, "Permuteattack: Counterfactual explanation of machine learning credit scorecards," 2020, \textit{arXiv:2008.10138}.

\bibitem{27}H. Zhao et al., "Remix: Towards the transferability of adversarial examples," \textit{Neural Netw.}, Vol. 163, pp. 367-378, Apr. 2023.


\bibitem{28}H. G. Jung et al., "Counterfactual explanation based on gradual construction for deep networks," \textit{Pattern Recognit.}, vol. 132, pp. 108-958, 2022.

\bibitem{29}C. H. Chang, E. Creager, A. Goldenberg, and D. Duvenaud,“Explaining image classifiers by counterfactual generation.” 2018,
\textit{arXiv:1807.08024.}

\bibitem{30}S. H. Na,  W. J. Nam, and S-W. Lee, "Toward practical and plausible counterfactual explanation through latent adjustment in disentangled space," \textit{Expert Syst. Appl.}, vol. 233, pp. 0957-4174, Dec. 2023.

\bibitem{31}Y. LeCun, "The MNIST database of handwritten digits," 1998.

\bibitem{32}Z. W. Liu et al., "Deep learning face attributes in the wild," in \textit{Proc. IEEE Int. Conf. Comput. Vis. (ICCV)}, pp. 3730-3738, 2015.

\bibitem{33}A. Akula, S. Wang, and S. C. Zhu, "Cocox: Generating conceptual and counterfactual explanations via fault-lines, " in \textit{Proc. AAAI Conf. Artif. Intell. (AAAI)}, pp. 2594–2601, 2020.













\bibitem{34} M. Scott and S.-I. Lee, "A unified approach to interpreting model predictions," in \textit{Proc. Int. Conf. Neural Inf. Process. Syst. (NeurIPS)}, pp. 4765–4774, 2017.

\bibitem{35} Q. Zheng et al., "Shap-CAM: Visual explanations for convolutional neural networks based on Shapley value," in \textit{Proc. Eur. Conf. Comput. Vis. (ECCV)}, pp. 459-474, 2022.








\bibitem{36}C. Wah, S. Branson, P. Welinder, P. Perona, and S. Belongie, "The caltech-UCSD birds-200-2011 dataset," \textit{Technical Report}, 2011.


\bibitem{37}E. Dataset, "Novel datasets for fine-grained image categorization," \textit{Proc. IEEE/CVF Conf. Comput. Vis. Pattern Recognit . (CVPR Workshop)}, 2011.



\bibitem{38} M. Sundararajan M, A. Taly, and Q. Yan," Axiomatic attribution for deep networks"in \textit{Proc. Int. Conf. Mach. Learn. (ICML)}, pp. 3319-3328, 2017.

\bibitem{39} J. T. Springenberg et al.," Striving for simplicity: The all convolutional net," 2014, \textit{arXiv:1412.6806}

\bibitem{40}V. Petsiuk, A. Das, and K. Saenko Kate," Rise: randomized input sampling for explanation of black-box models,"in \textit{Proc. Brit. Mach. Vis. Conf. (BMVC)}, pp. 151, 2018.


\bibitem{41}F. Ruth, P. Mandela, and V. Andrea,  "Understanding deep networks via extremal perturbations and smooth masks," in \textit{Proc. IEEE Int. Conf. Comput. Vis. (ICCV)}, pp. 2950–2958, 2019.

\bibitem{42}J. M. Kim et al., "Keep calm and improve visual feature attribution,"in \textit{Proc. IEEE Int. Conf. Comput. Vis. (ICCV)}, pp. 8350–8360, 2021.



\bibitem{43}M. Prabhushankar et al., " Contrastive explanations in neural networks," in \textit{Proc. IEEE Int. Conf. Image Process. (ICIP)}, pp. 3289-3293, 2020.


\bibitem{44}K. He, X. Zhang, S. Ren, and J. Sun, "Deep residual learning for image recognition," in \textit{Proc. IEEE/CVF Conf. Comput. Vis. Pattern Recognit. (CVPR)}, pp. 770-778, 2016.

\bibitem{45}K. Simonyan and A. Zisserman, "Very deep convolutional networks for large-scale image recognition," 2014, \textit{arXiv:1409.1556.}






\bibitem{46}W. Samek, A. Binder, G. Montavon, S. Lapuschkin, and K.-R. M{\"u}ller, "Evaluating the visualization of what a deep neural network has learned," \textit{IEEE Trans. Neural Netw. Learn. Syst.}, vol. 28, no. 11, pp. 2660--2673, 2016.









\end{thebibliography}
\end{document}